\documentclass[journal, twocolumn, 11pt]{IEEEtran}
\usepackage{float}
\usepackage{algorithm}
\usepackage{algpseudocode}
\usepackage{graphicx}
\usepackage{siunitx}
\usepackage{rotating,booktabs}
\usepackage{makecell}
\usepackage[inline]{enumitem}
\usepackage{amsmath,amssymb}
\usepackage{changepage}
\usepackage{textcomp,marvosym}
\usepackage{cite}
\usepackage{nameref,hyperref}
\usepackage[right]{lineno}
\usepackage[nopatch=eqnum]{microtype}
\DisableLigatures[f]{encoding = *, family = * }
\usepackage[table]{xcolor}
\usepackage{array}
\usepackage{tikz}

\definecolor{pink}{HTML}{fc5ba3}
\definecolor{navy}{HTML}{065a8f}
\definecolor{blue}{HTML}{6caff5}
\definecolor{red}{HTML}{eb484b}
\definecolor{light green}{HTML}{37d275}
\definecolor{lilac}{HTML}{CC99FF}
\definecolor{grey}{HTML}{4C4242}

\newcommand\submittedtext{
  \tiny{ © 2025 IEEE. This work has been submitted to the IEEE for possible publication. Copyright may be transferred without notice, after which this version may no longer be accessible.}
}

\newcommand\submittednotice{%
\begin{tikzpicture}[remember picture,overlay]
\node[anchor=south,yshift=10pt] at (current page.south) {\fbox{\parbox{\dimexpr0.65\textwidth-\fboxsep-\fboxrule\relax}{\submittedtext}}};
\end{tikzpicture}%
}

\begin{document}

\title{Exploring the impact of adaptive rewiring \\in Graph Neural Networks} 
\author{Charlotte~Cambier~van~Nooten \thanks{The authors are with the Data Science Department, Institute for Computing and Information Sciences, Radboud University, Nijmegen, The Netherlands. (Charlotte Cambier van Nooten and Christos Aronis contributed equally to this work. Yuliya Shapovalova and Lucia Cavallaro also contributed equally to this work.) \textit{(Corresponding author: lucia.cavallaro@ru.nl)}}, Christos~Aronis, Yuliya~Shapovalova, Lucia~Cavallaro}

\markboth{IEEE TRANSACTIONS ON NEURAL NETWORKS AND LEARNING SYSTEMS}{C. Cambier van Nooten, C. Aronis, 
\MakeLowercase{\textit{(et al.)}: 
Exploring the impact of adaptive rewiring in Graph Neural Networks}}

\maketitle

\submittednotice

\begin{abstract}
This paper explores sparsification methods as a form of regularization in Graph Neural Networks (GNNs) to address high memory usage and computational costs in large-scale graph applications. Using techniques from Network Science and Machine Learning, including \texorpdfstring{Erd\H{o}s--R\'enyi}{Erdos-Renyi} for model sparsification, we enhance the efficiency of GNNs for real-world applications. We demonstrate our approach on N-1 contingency assessment in electrical grids, a critical task for ensuring grid reliability. We apply our methods to three datasets of varying sizes, exploring Graph Convolutional Networks (GCN) and Graph Isomorphism Networks (GIN) with different degrees of sparsification and rewiring. Comparison across sparsification levels shows the potential of combining insights from both research fields to improve GNN performance and scalability. Our experiments highlight the importance of tuning sparsity parameters: while sparsity can improve generalization, excessive sparsity may hinder learning of complex patterns. Our adaptive rewiring approach, particularly when combined with early stopping, proves promising by allowing the model to adapt its connectivity structure during training. This research contributes to understanding how sparsity can be effectively leveraged in GNNs for critical applications like power grid reliability analysis.
\end{abstract}


\section{Introduction}
In recent years, Deep Learning has shown increased potential and diverse applications. However, this success is often tied to the increasing size and complexity of deep learning models, which require substantial computational resources and hinder scalability, particularly for large datasets and resource-constrained applications~\cite{thompson2020computational}. This computational burden presents a critical challenge: \textit{How can we maintain or even enhance model performance while reducing computational demands?}

A promising avenue to address this challenge is introducing sparsity into neural network architectures. By setting a portion of model weights to zero, we can substantially reduce memory footprint, improve generalization, mitigate overfitting, and enhance training and inference efficiency~\cite{frankle2018lottery,chen_unified_2021, han_deep_2016, gale2019state, hoefler2021sparsity, liu2023ten, liu2022unreasonable}.
Among sparsity-inducing techniques, Sparse Evolutionary Training (SET)~\cite{mocanu_scalable_2018} has emerged as a notable method that evolves the network's connectivity pattern during training to discover optimal sparse architectures. SET employs the \texorpdfstring{Erd\H{o}s--R\'enyi}{Erdos-Renyi} random graph model~\cite{peng_towards_2022, nowak2023fantastic} for initialization and applies a two-phase process at each epoch: first, a pruning phase where the smallest weight values are removed; then, a rewiring phase where removed connections are replaced at random positions with newly initialized weights, maintaining constant overall connectivity throughout training. We refer to this as \textit{fixed-rate rewiring}, where a fixed fraction $\zeta$ of weights is rewired each epoch. In contrast, we also explore \textit{adaptive rewiring} in this work, where $\zeta$ is adjusted during training.

It is important to distinguish rewiring from dropout~\cite{srivastava2014dropout}, though both serve as regularization techniques. Dropout randomly deactivates neurons during each forward pass, encouraging redundancy and robustness by preventing over-reliance on specific connections. In contrast, rewiring strategically removes low-magnitude connections and replaces them with new, randomly initialized ones, creating an evolving sparse topology throughout training. Unlike dropout's temporary deactivation, rewiring permanently modifies the network structure, enabling exploration of more efficient connectivity patterns while simultaneously acting as a regularizer.

Although sparsity and rewiring have been extensively studied in traditional neural networks, their application to GNNs poses unique challenges \cite{nesetril2014sparsity, ortega2018graph}. These challenges arise not from the sparse input graphs themselves, but from the interaction between model-level sparsity and message-passing dynamics in GNNs. Removing and reconfiguring connections can disrupt multi-hop information flow, reducing expressivity and worsening oversquashing—the bottleneck that occurs when information from distant nodes is compressed through message passing, and making training more difficult~\cite{alon2020bottleneck}. Consequently, rewiring for GNNs must carefully
balance sparsity, connectivity and generalisation.

Building on the principles of sparsity, this paper explores techniques to enhance GNN efficiency through sparsification and adaptive rewiring strategies. We develop an adaptive rewiring approach that dynamically adjusts the rewiring parameter $\zeta$ during training based on validation performance, offering flexible control over network connectivity. Additionally, we introduce a targeted early stopping mechanism for the rewiring process, allowing the model to discover and stabilize on optimal sparse architectures.

To validate our approach, we introduce N-1 contingency analysis in power grids as a novel application domain for GNN sparsification methods. This task predicts whether the grid remains stable after the failure of any single component (substation or power cable). The application is particularly well-suited for evaluating sparsification: power grids are naturally sparse graphs that require scalable, real-time analysis. Our work demonstrates how model sparsification can improve both computational efficiency and prediction performance for critical infrastructure applications.

To address these challenges and evaluate the efficacy of our proposed approach, we formulate the following research questions:
\begin{itemize}
    \item \textbf{Comparative analysis of sparsity techniques:} How does adaptive rewiring (our approach) compare to fixed-rate rewiring (SET-based) in terms of model performance, training stability, and final sparsity levels across different initial connectivity settings?
    \item \textbf{Sparsity tolerance and performance limits:} What is the relationship between sparsity (controlled by $\epsilon$) and model performance? At what sparsity levels does performance degradation become significant?
    \item \textbf{Real-world application impact:} How do our sparsification and rewiring methods impact both computational efficiency and prediction accuracy for N-1 contingency analysis in power grids?
\end{itemize}

Through these research questions, we aim to advance understanding of sparsity in GNNs and offer practical guidelines for developing efficient and scalable graph learning models. Our primary contributions are: \begin{enumerate*}[label=(\roman*)] \item a novel approach to GNN sparsity that integrates model weight sparsity with layer-wise adaptive rewiring; \item a comparative evaluation of our method against SET and dense GNN baselines (\textit{i.e.}, architectures with no sparsity nor rewiring); and \item a demonstration of the real-world efficacy of our approach in a critical infrastructure application. 
\end{enumerate*}

\section{Related work}

The concept of rewiring appears in various contexts throughout the literature, often referring to different mechanisms or objectives from those discussed in this paper. For instance, Gutteridge \textit{et al.}~\cite{gutteridge2023drew} employ graph rewiring to densify the input graph, thereby mitigating issues such as oversquashing~\cite{alon2020bottleneck, di2023does}. Oversquashing is the excessive compression of information from distant nodes, leading to substantial information loss. Such approaches enable multi-hop message passing, extending beyond local aggregation where information flows only between neighbouring nodes. Similar efforts to address oversquashing include the work of Barbero \textit{et al.}~\cite{barbero_locality-aware_2024}, who analyze the trade-offs between spatial~\cite{abboud2022shortest} and spectral rewiring~\cite{karhadkar2022fosr} techniques and propose a framework that better preserves locality in the input graph. For a broader overview of rewiring strategies aimed at mitigating oversquashing, we refer the reader to~\cite{attali2024rewiring}.

In contrast to these structural rewiring approaches, our use of rewiring is motivated by efficiency rather than expressivity. Specifically, we focus on achieving practical computational gains. While discovering sparse GNNs that outperform their dense counterparts is not our primary goal, we nonetheless explore the efficiency, the performance trade-off to ensure competitive results. Prior studies inspired by the lottery ticket hypothesis~\cite{frankle2018lottery}, such as Huang \textit{et al.}~\cite{huang2022you} and Chen \textit{et al.}~\cite{chen_unified_2021}, have demonstrated that sparse subnetworks can match or even surpass dense models. However, our emphasis lies in maintaining sparsity throughout training, known as sparse-to-sparse optimization, to promote computational efficiency. Related work by Peng \textit{et al.}~\cite{peng_towards_2022} compares the SET algorithm with other train-and-prune approaches, but unlike our study, it does not modify SET and focuses solely on benchmark datasets. In contrast, we evaluate our approach on practically relevant data, such as the N-1 contingency assessment in an electricity grid.

Finally, to avoid ambiguity, we clarify the distinction between rewiring and pruning. In rewiring, a random fraction $p$ of connections is removed (pruned) and replaced by an equal number of newly added random connections with randomly initialized weights. Since the total number of parameters remains constant, we do not consider rewiring as a pruning method. In this paper, pruning refers to any process that permanently removes weight connections. For simplicity, we use the term rewiring to denote the combined process of pruning and re-adding new random weights.

\section{Materials and methods}
This section develops our sparsification and rewiring framework for graph classification. We first formalize the graph classification problem and describe the GNN architectures used in our study. We then present classical sparsity techniques—Erd\H{o}s-R\'{e}nyi initialization and fixed-rate rewiring with a fixed $\zeta$ parameter. Finally, we introduce our key contribution: an adaptive rewiring algorithm that dynamically adjusts the rewiring intensity by adjusting the $\zeta$ parameter.

\subsection{Input graph classification}
We represent our data as an undirected graph $G = (V, E)$, where $V$ is the set of nodes (vertices) and $E$ is the set of edges (links) between the nodes. The graph structure, represented by the adjacency matrix $A$, encodes the relationships between the nodes. We apply our methods to datasets that naturally admit graph representations, including molecular structures and power grid topologies. Table~\ref{tab:notation} provides an overview of our notation. 

\begin{table*}[th]
    \centering
    \caption{Notation and description of a single graph.\label{tab:notation}}
    \begin{tabular}{l p{10cm}}
        \hline
        \textbf{Notation} & \textbf{Description} \\
        \hline
        \rowcolor{lightgray}\(|V|\) & Number of nodes \\
        \(|E|\) & Number of edges \\
        \rowcolor{lightgray}\(Y\) & Label \\
        \({F}^V\) & Input set of all node features, \({F}^V = \{F^V_1, . . . , F^V_{|V|}\}\) \\
        \rowcolor{lightgray}\({F}^E\) & Input set of all edge features, \({F}^E = \{F^E_1, . . . , F^E_{|E|}\}\) \\
        \(F^V_v \in \mathbb{R}^Q\) & \makecell[l]{Node feature matrix of \(v\)-th node}\\
        \rowcolor{lightgray}\(F^E_e \in \mathbb{R}^P\) & \makecell[l]{Edge feature matrix of \(e\)-th edge} \\ 
        \(\mathcal{N}(v)\) & Neighbouring nodes of node \(v\) \\
        \rowcolor{lightgray}\({X}\) & \makecell[l]{Input data, \({X}= \{G, {F}^V, {F}^E\}\), \\containing a single graph and the set of node and edge features} \\
        \hline 
    \end{tabular}
\end{table*}

For graph-level classification, the task is to learn a non-linear mapping $f$ from the input data to the output
\begin{equation}\label{eq:g}
    f: X \mapsto Y,
\end{equation}
where $X$ is a single graph with the set of features, and $Y$ is the graph-level label. The input data ($X$) contains graph structures ($\mathbb{G}$), node features ($F^V$) and edge features ($F^E$). 

We evaluate our approach on three datasets: MUTAG and PROTEINS (standard molecular graph benchmarks), and a power grid dataset for N-1 contingency analysis that predicts whether a grid configuration remains stable after any single component failure (see Sect.~\ref{subsec:datasets} for details).

\subsection{Graph Neural Network (GNN) Architectures}
 
Graph Neural Networks (GNNs)~\cite{scarselli2008graph, wu2020comprehensive} are a class of neural networks designed to operate on graph-structured data. A GNN can be defined as a function $f(\cdot)$ (see Eq.~\ref{eq:g}) that takes as input a graph $G=(V, E)$ and a set of node features ${F}^V = \{F^V_1, \dots, F^V_{|V|}\}$, edge features ${F} ^E = \{F^E_1, \dots , F^E_{|E|}\}$, and outputs the class prediction labels $\hat{y}$ for each graph. 

\begin{figure*}[th]
    \centering
    \includegraphics[width=1.0\linewidth]{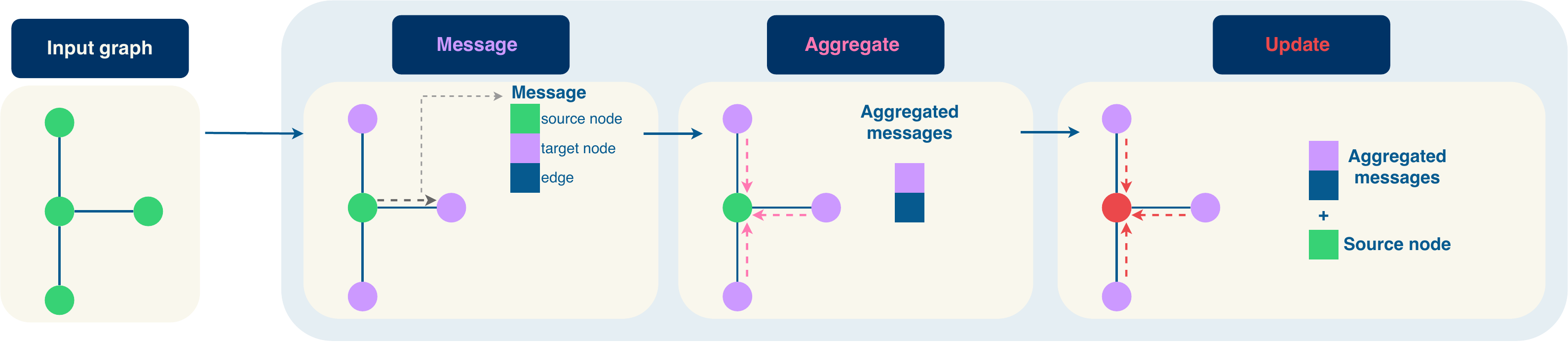}
    \caption{Information flow in a single step of message passing (MP) in a Graph Neural Network (GNN). The \textcolor{lilac}{\textsf{Message}} stage involves green \textit{source nodes} sending messages to lilac \textit{target nodes}. \textcolor{pink}{\textsf{Aggregate}} stage combines these messages from the lilac nodes. The \textcolor{red}{\textsf{Update}} stage integrates the aggregated messages into the source node's representation (which transitions to red during this phase), propagating information.
    }
    \label{fig:mpblock}
\end{figure*}

In GNNs, the output labels $\hat{y}$ for each graph are computed using a recursive message-passing (MP) procedure that operates on the local neighbourhood of nodes and edges. During message passing, a low-dimensional representation of the nodes, known as embeddings, is created. Each node receives information (\textit{i.e.}, features) from its neighbouring nodes and connected edges, updating its embedding using the features of the node itself and information from its neighbourhood in the graph, see Figure~\ref{fig:mpblock} for a schematic overview of the MP process in GNNs. This process is repeated for a fixed number of $K$ iterations, exploring the nodes $K$ hops away from the target node (the node for which the embedding is computed). The final node embeddings after $K$ iterations are used to compute label $y$~\cite{zhou_graph_2020}. 

GNN architectures differ primarily in their aggregation and update functions for computing node and edge embeddings. In this work, we used two different GNN architectures, a Graph Convolutional Network (GCN)~\cite{zhang2019graph} and a variation of the Graph Isomorphic Network (GIN)~\cite{xu_how_2019}, named GINE, proposed by Cambier van Nooten 
\textit{et al.}~\cite{cambier_van_nooten_graph_2025} that also includes edge features.
The choice of using these architectures is based on their efficiency in capturing graph structures, with GCNs utilising graph convolutions for local features and GINs employing isomorphism for richer node and edge representations. 

The original GIN architecture~\cite{xu_how_2019} does not incorporate edge features, only aggregating information from neighbouring nodes. However, incorporating edge features is important because it enables a richer representation of relationships between nodes, ultimately leading to better performance in tasks such as graph classification. GINE addresses this limitation by aggregating features from both neighbouring nodes and edges in each message-passing block, using Multi-Layer Perceptrons (MLPs) to update node and edge representations. A sequence of $K$ message-passing blocks enables information propagation across the $K$-hop neighbourhood, capturing both local node properties and larger graph structures~\cite{cambier_van_nooten_graph_2025}. Furthermore, while standard GNN architectures like the original GIN~\cite{xu_how_2019} use simple aggregation functions (mean, max, or sum), GINE employs MLPs as aggregation functions to enable more expressive representations and preserve complex node-edge dependencies. 

\subsubsection{Graph Convolutional Network (GCN) layers}\label{gcn} 
Graph Convolutional Networks (GCN)~\cite{kipf_semi-supervised_2017} update node features by aggregating the normalised features of their neighbours. This process enables GCNs to learn from the graph's structure for tasks such as node classification. The GCN initialises the node embeddings with $h_v^{(0)} = \sum_{u \in \mathcal{N}(v)}F_v^V$, and it does not use or update edge features.

To address the limitation of not using edge features, we introduce a variant called GCNE~\cite{cambier_van_nooten_graph_2025}, which incorporates edge features into the aggregation process. The adapted core update rule for GCNE is as follows:
\begin{equation*}
    g_e^{(k)} = W_{\text{edges}}^{(k)} \cdot g_e^{(k-1)},
\end{equation*}
\begin{equation*}
    a_v^{(k)} = \text{ReLU} \left(\sum_{u \in \mathcal{N}(v)} \left(\frac{1}{\sqrt{\hat{d}_v \hat{d}_u}} W_{\text{nodes}}^{(k)} h_u^{(k-1)} + g_{(v,u)}^{(k-1)}\right)\right).
\end{equation*}
Here, \(W_{\text{nodes}}^{(k)}\) represents a learnable weight matrix for the nodes at a given layer \(k\), \(W_{\text{edges}}^{(k)}\) is a learnable weight matrix for the edges, , while \(\hat{d}_{v}\) and \(\hat{d}_{u}\) are the normalized degrees for nodes \(v\) and \(u\), respectively. This adaptation enables the model to generate more nuanced representations of the graph data.

\subsubsection{Graph Isomorphic Network (GIN) layers}\label{app:gin}
The original Graph Isomorphism Network (GIN) framework does not incorporate edge features in its node update process~\cite{xu_how_2019}. Instead, node embeddings ($h_v^{(k)}$) are updated using a multi-layer perceptron (MLP) that combines the node's previous features with aggregated neighbour features. A GIN-inspired architecture, referred to as GINE~\cite{cambier_van_nooten_graph_2025}, was developed to include edge embeddings in the update process. In this architecture, embeddings for nodes and edges are first learnt from the raw features using MLPs. The embeddings $h_{v}^{(0)}$ are computed for each node, $v=\{1,\dots, |V|\}$ and the embeddings $g_{e}^{(0)}$ are computed for each edge, $e=\{1,\dots,|E|\}$:
\begin{equation*}
    h_{v}^{(0)} = \sum_{u \in \mathcal{N}(v)} \text{MLP}_1(F^V_u),
\end{equation*}
\begin{equation*}
    g_{e}^{(0)} = \text{MLP}_2(F^E_e).
\end{equation*}
Here, $MLP_1$ and $MLP_2$ are multi-layer perceptrons with non-linearity used for the embedding of node and edge features, $u \in \mathcal{N}(v)$ represents the neighbouring elements of target node $v$ and $F^E_{e}$ the edge features associated with edge $e$. The edge features are then updated within each message-passing block:
\begin{equation*}
    g_e^{(k)} = \text{MLP}_3 (g_e^{(k-1)}).
\end{equation*}
Here, $MLP_3$ is the multi-layer perceptron for the learnable edge features, and $g_{e}^{(k-1)}$ are the edge embeddings surrounding edge $e$. Next, node embeddings ($h_v^{(k)}$) are updated by aggregating both node and edge features ($a_v^{(k)}$ ) from the neighbourhood. To compute $a_v^{(k)}$ we take the target node $v$ and sum over all neighbouring nodes $u$ of $v$, $u \in \mathcal{N}(v)$. 
\begin{equation*}
    a_v^{(k)} = \sum_{u \in \mathcal{N}(v)} \text{ReLU} (h_u^{(k-1)}+g_{(v,u)}^{(k-1)}).
\end{equation*}
Finally, we combine the previous node embeddings $h_v^{(k-1)}$ with aggregated node embeddings $a_v^{(k)}$ to output the node embeddings of the current $k$-th MP-block:
\begin{equation*}
    h_v^{(k)} = \text{MLP}_4 ((1+\eta^{(k)})\cdot h_v^{(k-1)} + a_v^{(k)}).
\end{equation*}
Here, $MLP_4$ is the multi-layer perceptron for the learnable node features, and $\eta^{(k)}$ is a learnable parameter. After $K$ iterations of aggregation, the node representation captures the structural information within its $K$-hop network neighbourhood, defined by $K$ MP-blocks. The $K$-hop neighbourhood includes all nodes and edges within $K$ hops of the target node. In the following section, we focus on how adaptive sparsity is integrated specifically within these $MLP$ blocks, as this is where sparsity becomes an effective mechanism for controlling model complexity and enhancing interpretability.

After several iterations, the node embeddings capture structural information from their local neighbourhood. These are then aggregated to form a final graph representation ($h_G^{(k)}$) for classification. See Figure~\ref{fig:GNNblocks} for an overview of the different rules inside the MP-blocks.

\subsection{Employing Sparsity}
\begin{figure*}[ht]
    \centering
    \includegraphics[width=1.00\linewidth]{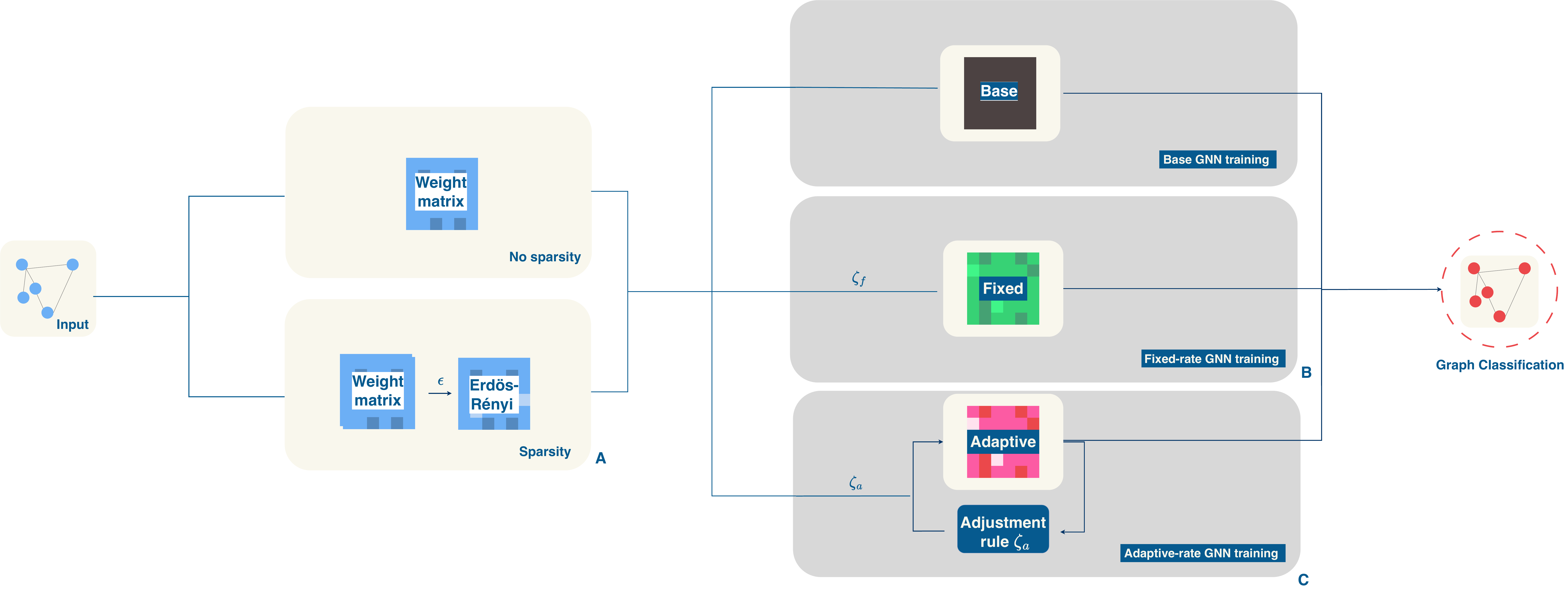}
    \caption{This figure illustrates a Graph Neural Network (GNN) framework incorporating sparsity for graph classification. The process starts with an input graph which is provided to a GNN. The weight matrices can be initialized using the \textcolor{blue}{\textsf{Erd\H{o}s-R\'{e}nyi}} method (option \textbf{\textsf{A}}) to introduce sparsity (\textit{i.e.}, $\epsilon$). If no sparsity is applied, the model proceeds with weight matrices.  Then during training, the GNN parameters are optimized via gradient based learning, while the sparse connectivity serves as a starting point for subsequent rewirings.  Sparsity is handled in two ways: a \textcolor{light green}{\textsf{Fixed}} (option \textbf{\textsf{B}}) approach maintains fixed sparsity (\textit{i.e.}, $\zeta_f$), while an \textcolor{pink}{\textsf{Adaptive}} (option \textbf{\textsf{C}}) approach modifies the sparsity pattern using an adjustment rule (\textit{i.e.}, $\zeta_a$) throughout training. If neither fixed-rate nor adaptive rewiring is selected, the model follows the \textcolor{grey}{\textsf{Base}} training mode, as shown in the upper branch of the diagram. The output of the GNN is a graph classification.}
    \label{fig:sparse_overview}
\end{figure*}

\begin{figure*}[!ht]
    \centering
    \includegraphics[width=0.9\linewidth]{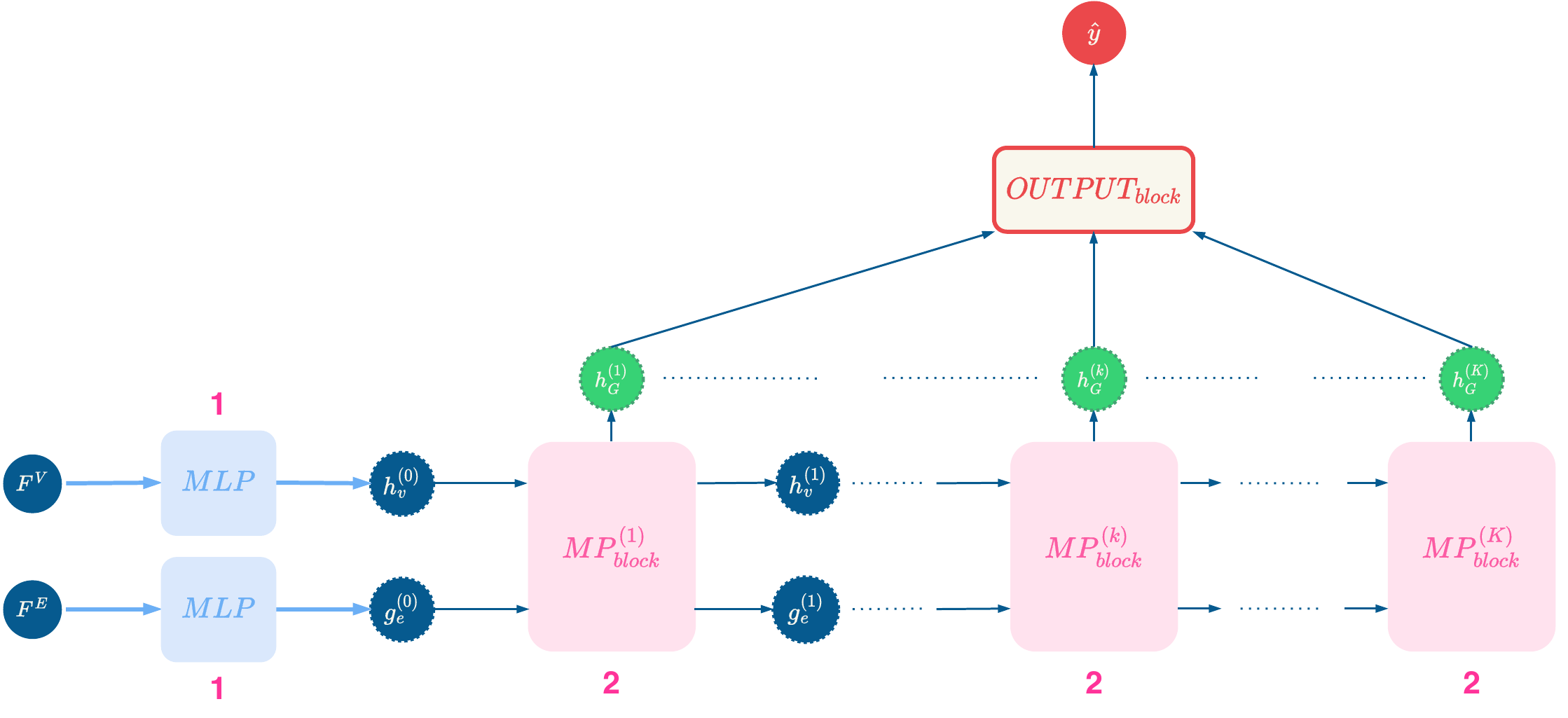}
    \caption{The diagram where \textit{sparsity} (\textcolor{pink}{\textbf{\textsf{1}}}) and \textit{rewiring} (\textcolor{pink}{\textbf{\textsf{2}}}) is applied to the GNN. Here, the \texorpdfstring{Erd\H{o}s--R\'enyi}{Erdos-Renyi} mask with sparsity level ($\epsilon$) is applied directly to the model’s weight matrices ($MLP$), reducing the number of trainable connections before training begins. Rewiring is introduced during the message-passing (\textit{MP}) process. Rewiring ($\zeta$) updates the model connectivity during training by modifying the weights $W_{edges}$/$W_{nodes}$ in GCN or $MLP_3$/$MLP_4$ in GINE. After the MP blocks, the \textsf{OUTPUT} layer aggregates the learned node and edge embeddings to produce a final graph-level representation via a linear layer (GCN) or $MLP_{out}$ (GINE).}
    \label{fig:GNNblocks}
\end{figure*}

To comprehensively investigate the impact of sparsity on GNN performance, we employ a three-way approach: \begin{enumerate*}[label=(\roman*)] \item controlled sparsity using the Erd\H{o}s-R\'{e}nyi model ($\epsilon$)~\cite{mocanu_scalable_2018}, \item fixed layer-wise rewiring during training with a fixed parameter ($\zeta_f$), and \item our adaptive layer-wise rewiring during training with an adaptable parameter ($\zeta_a$), see Sect.~\ref{sec:dynamic}.\end{enumerate*} We explore different combinations of these techniques to differentiate their effects and understand the benefits of fixed-rate versus adaptive-rate rewiring. Figure~\ref{fig:sparse_overview} presents a basic overview of the framework, using three different sparsity applications. Figure~\ref{fig:GNNblocks} presents the detailed overview of rewiring locations inside the GNNs.

Throughout training, we maintain constant sparsity levels for three reasons. First, constant sparsity keeps models compact, reducing memory and computational costs. Second, it enables systematic analysis of how sparsity affects training dynamics—without this constraint, models might gradually restore pruned connections, negating sparsification benefits. Third, in sparse models, there are far fewer connections available, so the selective rewiring of those connections can have a significant impact on the model's performance. Rewiring enables a sparse model to adaptively reallocate its limited connections, potentially improving overall performance. In contrast, dense models (\textit{i.e.}, $\epsilon=0$, $\zeta=0$) already have redundant connections, so the benefits of rewiring might be less impactful. 

\subsubsection{Model sparsity with 
\texorpdfstring{Erd\H{o}s--R\'enyi ($\epsilon$)}{Erdos-Renyi (epsilon)}} 

We introduce sparsity by applying the \texorpdfstring{Erd\H{o}s--R\'enyi}{Erdos-Renyi} random graph model $G(n,p)$ to the GNN's weight matrices. In this context, neurons correspond to nodes and weight connections correspond to edges in the random graph. Each connection (edge) between neurons exists independently with probability $\epsilon$. 

For a weight matrix $\mathbf{W} \in \mathbb{R}^{N \times M}$, we generate a binary mask $\mathbf{B}$ where $B_{ij} \sim \text{Bernoulli}(\epsilon)$, yielding the sparse weight matrix $W_{\text{sparse}, ij} = W_{ij} \times B_{ij}$. Here, $W_{ij}$ represents the weight connecting input neuron $i$ to output neuron $j$, and $\epsilon \in [0, 1]$ controls the connection probability. This results in sparsity level $S = 1 - \epsilon$ (fraction of zero weights) and an expected $\mathbb{E}[\|\mathbf{W}\|_0] = \epsilon \times N \times M$ non-zero parameters, where $\|\mathbf{W}\|_0$ denotes the $L_0$ norm.

We systematically explore the impact of sparsity on model performance using $\epsilon \in \{0.1, 0.3, 0.5, 0.7\}$, examining how different connectivity levels interact with subsequent rewiring strategies (see Sect.~\ref{subsect:exp_setup}).

\subsubsection{\texorpdfstring{Fixed-rate rewiring ($\zeta_f$) during training}{Fixed-rate rewiring (zeta\_f) during training}} 
We implement fixed-rate rewiring following the SET approach~\cite{mocanu_scalable_2018}, where the rewiring parameter $\zeta_f$ remains fixed throughout training. At each rewiring step, $\zeta_f$ determines the proportion of connections ($C$) to rewire in each layer ($l$), maintaining a constant rewiring intensity across all epochs.
In fixed-rate rewiring, the goal is to adapt the sparsity of the MP-blocks (within the GNN layers) throughout training, which happens in each epoch. See Algorithms~\ref {alg:static_rewiring} and \ref{alg:helpers} for a more detailed overview of the fixed-rate rewiring application. To achieve this, we first quantify the importance of each connection based on the value of its weight. In particular, connections are ranked in ascending order by absolute weight magnitude to identify the least important ones for removal. The fixed-rateparameter, $\zeta_f \in [0, 1]$, is then used to determine the proportion of connections to be rewired in each layer at each rewiring step; specifically, we rewire the least important $\lfloor \zeta_f \cdot |C^{(l)}| \rfloor$ connections in layer $l$, experimenting with $\zeta_f$ values in the range of 0.1, 0.3, 0.5 and 0.7 to investigate the effect of different fixed levels of rewiring intensity (See Sect.~\ref{subsect:exp_setup}). The rewiring process involves removing the $\lfloor \zeta_f \cdot |C^{(l)}| \rfloor$ least important connections from $C^{(l)}$ and adding an equal number of new connections using random reconnection. This approach involves computing the mean $\mu$ and the standard deviation $\sigma$ of the remaining weights, and then sampling the new weights from a normal distribution $N(\mu, \sigma^{2})$ to ensure that they match the existing weight distribution. By keeping the $\zeta_f$ parameter constant throughout the training, we can examine the effect of a consistent level of sparsity adaptation on the training and performance of GNN. 

\begin{algorithm*}
\caption{Fixed-rate rewiring algorithm}
\label{alg:static_rewiring}

\begin{algorithmic}[1]
\Require {$\text{GNN}$: Graph Neural Network model; $l$: Current layer in the $\text{GNN}$; $\zeta_f$: Fixed-rate rewiring parameter.} 

\Statex
\Function{FixedRewiring}{$\text{GNN}$, $l$, $\zeta_f$}
    \State $C^{(l)} \gets $\text{GNN}$.\Call{getConnections}{l}$ \Comment{Get connections in layer $l$}

    \State $\text{rankedConnections} \gets \Call{sort}{C^{(l)}}$ 
    \State $\text{numConnectionsRewire} \gets \Call{floor}{(\zeta_f \cdot |C^{(l)}|)}$

    \State GNN.\Call{remove}{rankedConnections, numConnectionsRewire} \Comment{Pruning}
    \State GNN.\Call{addRandomRewired}{numConnectionsRewire} \Comment{Rewiring}

    \State \Return $\text{GNN}$ \Comment{GNN with updated weights}
\EndFunction
\end{algorithmic}
\end{algorithm*}

\begin{algorithm*}
\caption{Helping functions}
\label{alg:helpers}

\begin{algorithmic}[1]

\Statex
\Function{remove}{rankedConnections, numConnectionsRewire} 
    \State $\text{connectionsToRemove} \gets \text{rankedConnections}[1:\text{numConnectionsRewire}]$
    \For{each $c$ in connectionsToRemove}
        \State $\text{GNN}.\Call{removeConnection}{l, c}$
    \EndFor
\EndFunction
\Statex
\Function{addRandomRewired}{numConnectionsRewire} 
    \State $w' \gets \text{GNN}.\Call{getRemainingWeights}{l}$
    \State $\mu \gets \Call{mean}{w'}$
    \State $\sigma \gets \Call{standardDeviation}{w'}$
    \For{$i \gets 1$ to $\text{numConnectionsRewire}$}
        \State $c^* \gets \Call{generateRandomConnections}{\text{GNN}, l}$
        \State $w^* \gets \Call{sampleFromNormalDistribution}{\mu, \sigma}$
        \State $\text{GNN}.\Call{addConnections}{l, c^*, w^*}$
    \EndFor 
\EndFunction
\Statex

\end{algorithmic}
\end{algorithm*}

\subsection{Proposed adaptive rewiring \texorpdfstring{($\zeta_a$)}{zeta\_a} during training} \label{sec:dynamic}

In contrast to fixed-rate rewiring with fixed $\zeta_f$, adaptive rewiring employs a parameter $\zeta_a$ that is dynamically adjusted during training based on validation performance. The rewiring mechanism itself—magnitude-based connection removal and random reconnection—remains identical to fixed-rate rewiring, but the intensity adapts to the model's learning progress (see Algorithms~\ref{alg:dynamic_rewiring} and \ref{alg:helpers}).

\textbf{Adaptive mechanism.} We regulate $\zeta_a$ using a sliding window of size $q$ that maintains a history of recent validation accuracies. At each rewiring step, we compute the average accuracy over this window and adjust $\zeta_a$ accordingly: if current accuracy exceeds the window average (indicating effective connectivity), we decrease $\zeta_a$ to promote stability; if current accuracy falls below average (suggesting suboptimal structure), we increase $\zeta_a$ to encourage exploration. The parameter is constrained within bounds $[\zeta_{min}, \zeta_{max}]$ throughout training.

\textbf{Early stopping for rewiring.} We incorporate a targeted early stopping mechanism that halts rewiring while allowing training to continue. This mechanism monitors validation loss with patience parameter $p$ and minimum improvement threshold $\text{min}_\Delta$. When validation loss fails to improve by at least $\text{min}_\Delta$ for $p$ consecutive epochs, rewiring stops, preventing unnecessary architectural changes once an effective sparse structure is found. This approach helps prevent overfitting, reduces computational cost, and improves training stability.

\textbf{Experimental setup for effects of $\zeta_a$.}
We experiment with $\zeta_a$ values ranging from 0.05 to 0.7 (increments of 0.05) to explore the impact of different adaptive rewiring intensities. Adaptive rewiring is applied at the end of each epoch to each GNN layer, rewiring $\lfloor \zeta_a \cdot |C^{(l)}| \rfloor$ connections per layer based on the current $\zeta_a$ value.

By comparing fixed-rate rewiring (fixed $\zeta_f$) with adaptive rewiring (dynamic $\zeta_a$ plus early stopping), we assess the benefits of adaptive sparsity management during training and investigate how these strategies interact with sparsity controlled by \texorpdfstring{Erd\H{o}s--R\'enyi}{Erdos-Renyi} parameter $\epsilon$. 

\begin{algorithm*}
\caption{Adaptive rewiring with early stopping}
\label{alg:dynamic_rewiring}
\begin{algorithmic}[1]
\Require {$\text{GNN}$: GNN Model; $l$: Layer; $\zeta_{min}, \zeta_{max}$: Zeta bounds; $q$: Window size; $\delta$: Early stop patience; $\text{min}_\Delta$: Early stop min delta.}
\Statex
\Function{AdaptiveRewiring}{$\text{GNN}$, $l$, $\zeta_{min}$, $\zeta_{max}$, $q$, $\delta$,  $\zeta_a$, $\text{min}_\Delta$}
    \State $\text{accuracyHistory} \gets []$
    \State $\text{bestLoss} \gets \infty$
    \State $i \gets 0$ \Comment{Counter}
    \State $\text{Rewiring} \gets \text{TRUE}$

    \While{$\text{Rewiring}$}
        \State $C^{(l)} \gets \text{GNN}.\Call{getConnections}{l}$ \Comment{Get connections in layer $l$}
        
        \State $\text{rankedConnections} \gets \Call{sort}{C^{(l)}}$ 
        \State $\text{numConnectionsRewire} \gets \Call{floor}{(\zeta_a \cdot |C^{(l)}|)}$

        \State GNN.\Call{remove}{rankedConnections, numConnectionsRewire} \Comment{Pruning}
        \State GNN.\Call{addRandomRewired}{numConnectionsRewire} \Comment{Rewiring}

        \State $\text{accuracy} \gets \Call{evaluateValidationAccuracy}{\text{GNN}}$ \Comment{Sliding window}
        \State $\Call{\text{accuracyHistory}.append}{\text{accuracy}}$

        \If{$|\text{accuracyHistory}| > q$} \Comment{Performance adjustment}
            \State $\text{averageAccuracy} \gets \Call{avg}{\text{accuracyHistory}[|\text{accuracyHistory}| - q :]}$
            \If{$\text{accuracy} > \text{averageAccuracy}$}
                \State $\zeta_a \gets \Call{max}{(\zeta_{min}, \zeta_a - 0.05)}$ \Comment{Reduce $\zeta_a$}
            \Else
                \State $\zeta_a \gets \Call{min}{(\zeta_{max}, \zeta_a + 0.05)}$ \Comment{Increase $\zeta_a$}
            \EndIf
        \EndIf
        \State $\text{loss} \gets \Call{evaluateValidationLoss}{\text{GNN}}$ 
        
        \State $\text{Early stopping (}\zeta_a\text{)}$
    \EndWhile

    \State \Return $\text{GNN}$ \Comment{GNN with updated weights}
      
\EndFunction
\end{algorithmic}
\end{algorithm*}

\section{Experiments}

\subsection{Datasets} 
\label{subsec:datasets}
In this study, we used two widely known graph classification datasets, MUTAG and PROTEINS~\footnote{Both MUTAG and PROTEINS datasets are publicly available from \url{https://chrsmrrs.github.io/datasets/docs/datasets/}} ~\cite{morris_tudataset_2020} and a dataset for power grids, which we call \textit{N-1 dataset}, to evaluate the performance of our proposed GNN architecture. By choosing datasets from diverse domains, we aim to evaluate the proposed GNN across diverse application domains, thus advancing our central research question. In this section, we provide a detailed description of the datasets. Table~\ref{tab:sum_data} summarises their graph characteristics.

\paragraph{MUTAG dataset} The MUTAG dataset, a standard benchmark in chemical informatics, consists of chemical compounds represented as graphs. Each node encodes atom type (\textit{e.g.}, carbon, nitrogen, oxygen), and each edge indicates bond type (\textit{e.g.}, single, double, aromatic). The classification task is to predict the mutagenic effect of these compounds. Its relatively small size and intricate graph structures test the model's ability to capture subtle patterns, such as the presence of aromatic rings and nitro groups, which are significant motifs for assessing mutagenicity. Identifying these chemical structures is crucial for evaluating the effectiveness of our GNN in learning meaningful graph-level representations.

\paragraph{PROTEINS dataset} The PROTEINS dataset, a standard reference from protein structure databases, presents a more complex and larger classification challenge. It includes node features representing amino acid properties and edge features denoting structural proximities. In contrast to the small molecular graphs in MUTAG, PROTEINS contains graphs that are significantly larger and more heterogeneous, with an average of over 39 nodes per graph and richer structural variability. Each node encodes biochemical properties of amino acids, while edges capture 3D spatial proximities rather than simple chemical bonds, resulting in more diverse and less regular connectivity patterns. These characteristics make the task of classifying proteins based on structural similarity notably more demanding, requiring GNNs to integrate information across wider neighbourhoods and handle greater variability in graph size, topology, and feature distributions.

\paragraph{N-1 dataset} For the power grid dataset, we formulate the problem as a graph classification task to assess power grid reliability under N-1 contingency conditions. Given an input graph, the objective is to predict whether the grid remains stable after the failure (removal) of any single component~\cite{cambier_van_nooten_graph_2025}. We train a GNN to classify grid configurations as 'stable' or 'unstable.' The dataset from Distribution System Operation (DSO) Alliander features four Dutch grid structures under various load conditions ~\cite{cambier_van_nooten_graph_2025} \footnote{Currently, this dataset is available upon request.}. Each sample includes graph features labeled by the satisfaction of the N-1 criterion, which determines the reliability of the grid under single-component failure conditions (see Table~\ref{tab:feat_table}). The dataset presents a class distribution of 50\% 'stable' and 50\% 'unstable' configurations, providing a balanced view of grid stability scenarios.

\begin{table}[t]
    \centering
    \small 
    \caption{Summary of the three datasets: MUTAG, PROTEINS and N-1.\label{tab:sum_data}}
    \begin{tabular}{l|c|c|c} \hline
        \textbf{Dataset} & \textbf{Graphs} & \textbf{Nodes} & \textbf{Edges}\\ \hline
        \rowcolor{lightgray}MUTAG & 118 & 17.9 & 19.8 \\
        PROTEINS & 1 113 & 39.1 & 72.8 \\
        \rowcolor{lightgray}N-1 & 64 000 & 350.5 & 393.3 \\ \hline
    \end{tabular}
\end{table}

\begin{table*}[th]
    \centering
    \caption{Node and edge features used for N-1 graph classification with GNNs. The physical units include Ampere (A), Watt (W), Volt (V), and Ohm ($\Omega$).
    \label{tab:feat_table}}
    \begin{tabular}{l|l}
    \hline
    \textbf{Node features (stations)}            & \textbf{Edge features (cables)  }                                                                 \\ \hline
    Power consumption (W) & Impedance ($\Omega$) \\ (or generation on load of node)                 & Nominal current (if NOP) (A)    \\
    Voltage of (radial) initial state (V) & Normal Open Point (NOP) \\
    Voltage of closed state (V) & Current of initial (radial) state (A)                                                                          \\
    Node degree                         & Current of closed state (A)                                                                    \\ \hline 
    \end{tabular}
\end{table*}

\subsection{Experimental setup}
\label{subsect:exp_setup}
To evaluate the impact of sparsity and rewiring on model performance, we performed a set of experiments using variations of standard GCN, GCN with edge features, standard GIN, and GIN with edge features. Our experiments were structured around varying values of sparsity ($\epsilon$) and rewiring rate ($\zeta$). To ensure stability and reduce sensitivity to randomness, the dataset was split into training, validation, and test sets with an 80:10:10 ratio, and results were averaged over 50 runs (each run with a different seed for randomization).

The experiments evaluated all combinations of $\epsilon$ and $\zeta$ (i.e., both fixed-rate $\zeta_f$ and adaptive-rate $\zeta_a$ rewiring) to assess the impact of sparsity and rewiring. The values for $\epsilon$, controlling the sparsity, were 0, 0.1, 0.3, 0.5, 0.7, and 0.9. The values for $\zeta$, controlling the rewiring, were 0, 0.1, 0.3, 0.5, and 0.7. We chose these values to capture the full spectrum from no sparsity or rewiring to high levels of sparsity and rewiring. 

We first assess the isolated effect of sparsity by setting $\zeta=0$ and evaluating the model's performance for different $\epsilon$ values. This reveals how sparse connectivity alone impacts learning. Next, we examine the effect of fixed-rate rewiring by setting each value $\zeta_f$ and averaging across $\epsilon$. We use a similar procedure for adaptive rewiring ($\zeta_a$). For both fixed-rate and adaptive-rate rewiring, we experiment with and without early stopping to see its effect on training stability and model generalisation.

\subsection{Training details}
All experiments utilize PyTorch 1.13.1 and PyTorch Geometric, incorporating the GCN and GIN frameworks. We use three datasets: MUTAG, PROTEINS, and N-1. The models are trained and evaluated on 2 NVIDIA RTX A6000 GPU nodes. 

We trained our GNN frameworks\footnote{The code is available at \url{https://github.com/christosaronis/RewireGNNs}.} with a set of hyperparameters (number of MP-blocks, activation function, learning rate, batch size, hidden size of the $MLPs$). The baseline GNN frameworks (without sparsity or network rewiring) use dropout. Dropout is used in neural networks to randomly deactivate some weights in each forward pass. This acts as a regularizer, helping to improve model generalization. Rewiring, on the other hand, involves modifying the network connectivity to achieve enhanced performance and efficiency by finding an optimal set of connections. The optimizer minimizes the cross-entropy loss. After hyperparameter optimization, the best accuracy is achieved using early stopping.

One-dimensional batch normalization is applied after each MP block. The parameter $\epsilon^{(k)}$ in GINE is updated by gradient descent. 
The embeddings are created using two-layer $MLP$s for the individual network configurations. We use the Adam optimiser with a learning rate of $1 \times 10^{-4}$ for GNN training. As mentioned in Sect.~\ref {subsect:exp_setup}, we used different values for $\epsilon$ and $\zeta$ and compared the performance of the various settings. To evaluate the impact of these configurations, we report on metrics including accuracy. The results are averaged across the 50 runs to ensure robustness. This approach enables a comprehensive evaluation of each configuration's performance, facilitating a clear comparison of the impact of sparsity and rewiring.

\section{Results} 
This section presents an analysis of our experimental findings, focusing on the impact of model sparsity and of fixed-rate $\zeta_{f}$ and adaptive rewiring  $\zeta_{a}$ on the performance of GNNs across various datasets, including MUTAG, PROTEINS, and the N-1 contingency analysis dataset. To ensure stability and reduce sensitivity to randomness, we conducted each experiment with multiple runs using different random seeds and evaluated our results using standard accuracy metrics.

The analysis examines the impact of model sparsity and rewiring strategies on the performance of GNNs, focusing on the MUTAG and PROTEINS datasets. The N-1 contingency dataset is covered in Sect.~\ref{sec:use_case} as a case study for large datasets. We evaluated sparsity techniques, trade-offs between accuracy and computational efficiency, and practical implications. 

We begin by examining the effects of sparsity, followed by an analysis of fixed-rate and adaptive rewiring strategies. We then explore the combined impact of these sparsity parameters and dive into the specific results obtained from our N-1 case study. Finally, we provide insights into how sparsity affects the number of model parameters, highlighting the potential for more efficient GNN models.

\subsection{The effects of sparsity \texorpdfstring{($\epsilon$)}{epsilon}}
To assess the impact of sparsity, we evaluated the performance of GNN models on the MUTAG and PROTEINS datasets across a range of $\epsilon$ values.

In Table~\ref{tab:comparison_epsilon}, we can observe that on the MUTAG dataset, both the GCNE and GINE models show improved performance with moderate levels of sparsity. Specifically, GCNE achieves its highest test accuracy at 0.75 with $\epsilon = 0.1$, while GINE peaks at 0.89 when $\epsilon = 0.3$. This suggests that a degree of sparsity can act as a regularizer. However, as expected, excessive sparsity (\textit{i.e.}, $\epsilon = 0.9$) results in a performance drop for both models.

\begin{table*}[ht]
    \centering
    \caption{Effect of sparsity ($\epsilon$) on GNN model test accuracy. The results are averaged over 50 runs (different amounts of pruning), and the best results are highlighted in bold.\label{tab:comparison_epsilon}}
    \begin{tabular}{l|l|c|c|c|c|c|c}
    \hline
    \multicolumn{1}{l|}{\textbf{Dataset}} & \multicolumn{1}{l|}{\textbf{Model}}   & \multicolumn{5}{c}{\textbf{Sparsity $\epsilon$}}  \\ \cline{3-8}
    &  & \textbf{0} & \textbf{0.1} & \textbf{0.3} & \textbf{0.5} & \textbf{0.7} & \textbf{0.9} \\
    \hline
    \rowcolor{lightgray} MUTAG & GCNE & 0.64 & \textbf{0.75} & 0.72 & 0.70 &     0.68 & 0.62 \\
    MUTAG & GINE & 0.65 & 0.81 & \textbf{0.89} & 0.85 & 0.79 & 0.71 \\
    \rowcolor{lightgray} PROTEINS & GCN & 0.68 & 0.71 & 0.73 & \textbf{0.75}     & 0.72 & 0.65 \\
    PROTEINS & GIN & 0.78 & \textbf{0.82} & 0.79 & 0.76 & 0.73 & 0.70\\
    \hline
    \end{tabular}
\end{table*}

The PROTEINS dataset displays a similar trend for GCN, with the highest accuracy of 0.75 observed at $\epsilon = 0.5$. GIN, on the other hand, shows a gradual increase in performance to $\epsilon = 0.1$, reaching a maximum of 0.82. Again, extreme sparsity leads to a reduction in accuracy for GCN.

The results indicate that an optimal level of sparsity exists for each model and dataset. Introducing no sparsity may fail to provide the needed regularisation, while excessive sparsity can hinder the model's ability to learn meaningful representations. The optimal value $\epsilon$ appears to depend on both the dataset and the model, which requires careful tuning.

\subsection{The effects of fixed-rate rewiring \texorpdfstring{($\zeta_f$)}{zeta\_s)}}
To show the effect of fixed-rate rewiring, we averaged the performance across various sparsity levels $\epsilon$. Table~\ref{tab:comparison_zetas} summarises the performance achieved in both fixed-rate ($\zeta_f$) and adaptive ($\zeta_a$) rewiring, varying GNN architectures, and datasets.

On the MUTAG dataset, GCNE achieves its best performance, which is 0.74, at $\zeta_f=0.1$, indicating an optimal level of fixed-rate rewiring. GINE, on the contrary, shows the best result, reaching up to 0.88, at $\zeta_f=0.1$. Beyond $\zeta_f=0.1$, we observed gradual performance degradation in both models, suggesting that excessive fixed-rate rewiring during training hinders representational capacities.

Similarly, on the PROTEINS dataset, GCN achieves its highest accuracy of 0.75 at $\zeta_f=0.1$. GIN, on the contrary, shows the best performance (\textit{i.e.}, 0.83) also at $\zeta_f=0.1$. Again, higher values of $\zeta_f$ lead to a decrease in accuracy.

These results suggest that moderate fixed-rate rewiring ($\zeta_f=0.1$) can be beneficial for GNN models, potentially acting as an additional regularizer and improving generalisation. However, excessive fixed-rate rewiring ($\zeta_f > 0.1$) appears to negatively impact performance, possibly due to the model's inability to learn complex patterns with a significantly changed connectivity structure. 
a
Similarly, Table \ref{tab:comparison_zetas} shows that rewiring also acts as a regularizer. The test accuracy peaks at an intermediate value of $\zeta$, indicating that both insufficient and excessive rewiring can be detrimental to performance.

\subsection{The effects of adaptive rewiring \texorpdfstring{$\zeta_a$}{zeta\_a} with early stopping}
To assess the impact of adaptive rewiring, we evaluated model performance on the datasets while adjusting the sparsity level during training. It is important to note that the values of adaptive rewiring shown on table \ref{tab:comparison_zetas} correspond to the initial rewiring intensities around which the adaptive mechanism adjust during training. 

\begin{table*}[tb]
    \centering
    \caption{Effect of rewiring ($\zeta_f$ and $\zeta_a$) on GNN model test accuracy. The results are averaged over 50 runs, and the best results per rewiring strategy are underlined, and overall over both strategies are highlighted in bold. For each $\zeta$ setting, results are averaged across sparsity levels of $\epsilon$ = $0.1$, $0.3$, $0.5$, and $0.7$. An important distinction between fixed-rate and adaptive-rate rewiring is that the values of $\zeta_{a}$ correspond to the initial rewiring intensities around which the adaptive mechanism adjust during training.
    \label{tab:comparison_zetas}}
    \begin{tabular}{l|l|c|c|c|c|c|c|c|c|c|c}
    \hline
    \multicolumn{1}{l|}{\textbf{Dataset}} & \multicolumn{1}{l|}{\textbf{Model}}   & \multicolumn{5}{c|}{\textbf{Fixed rewiring $\zeta_f$}} & \multicolumn{5}{c}{\textbf{Adaptive rewiring $\zeta_a$}}  \\
    \cline{3-12}
    &  & \textbf{0} & \textbf{0.1} & \textbf{0.3} & \textbf{0.5} & \textbf{0.7} & \textbf{0} & \textbf{0.1} & \textbf{0.3} & \textbf{0.5} & \textbf{0.7} \\
    \hline
    \rowcolor{lightgray} MUTAG & GCNE & 0.70 & \underline{0.74} & 0.71 & 0.68 & 0.65 & 0.72 & \textbf{\underline{0.76}} & 0.73 & 0.71 & 0.69  \\
    MUTAG & GINE & 0.81& \underline{0.88} & 0.85 & 0.81& 0.77 & 0.83 & \textbf{\underline{0.90}} & 0.87 & 0.85 & 0.82 \\
    \rowcolor{lightgray} PROTEINS & GCN & 0.71 & \underline{0.75} & 0.73  & 0.70 & 0.68 & 0.73 & \textbf{\underline{0.77}} & 0.75 & 0.72 & 0.70 \\
    PROTEINS & GIN  & 0.79 & \underline{0.83} & 0.80 & 0.77 & 0.75 & 0.81 & 0.85 & \textbf{\underline{0.87} }& 0.83 & 0.79 \\ 
    \hline
    \end{tabular}
\end{table*}

On the MUTAG dataset, GCNE achieves its best performance of 0.76 at $\zeta_a=0.1$, while GINE obtains its highest accuracy (0.90) at the same sparsity level. The performance of both models tends to decrease as $\zeta_a$ increases, suggesting that moderate adaptive rewiring, particularly around $\zeta_a=0.1$, is beneficial for this dataset.

For the PROTEINS dataset, GCN reaches its peak performance of 0.77 at $\zeta_a=0.1$. GIN, on the other hand, shows its best results (0.87) at a slightly higher rate, $\zeta_a=0.3$. Similarly to the MUTAG dataset, higher $\zeta_a$ values generally lead to a decrease in accuracy for both models.

These results suggest that adaptive rewiring, when combined with early stopping, can effectively adjust the sparsity level during training, resulting in improved performance.

\subsection{Fixed versus adaptive rewiring}
\begin{figure*}[tb]
    \centering
    \includegraphics[width=0.55\linewidth]{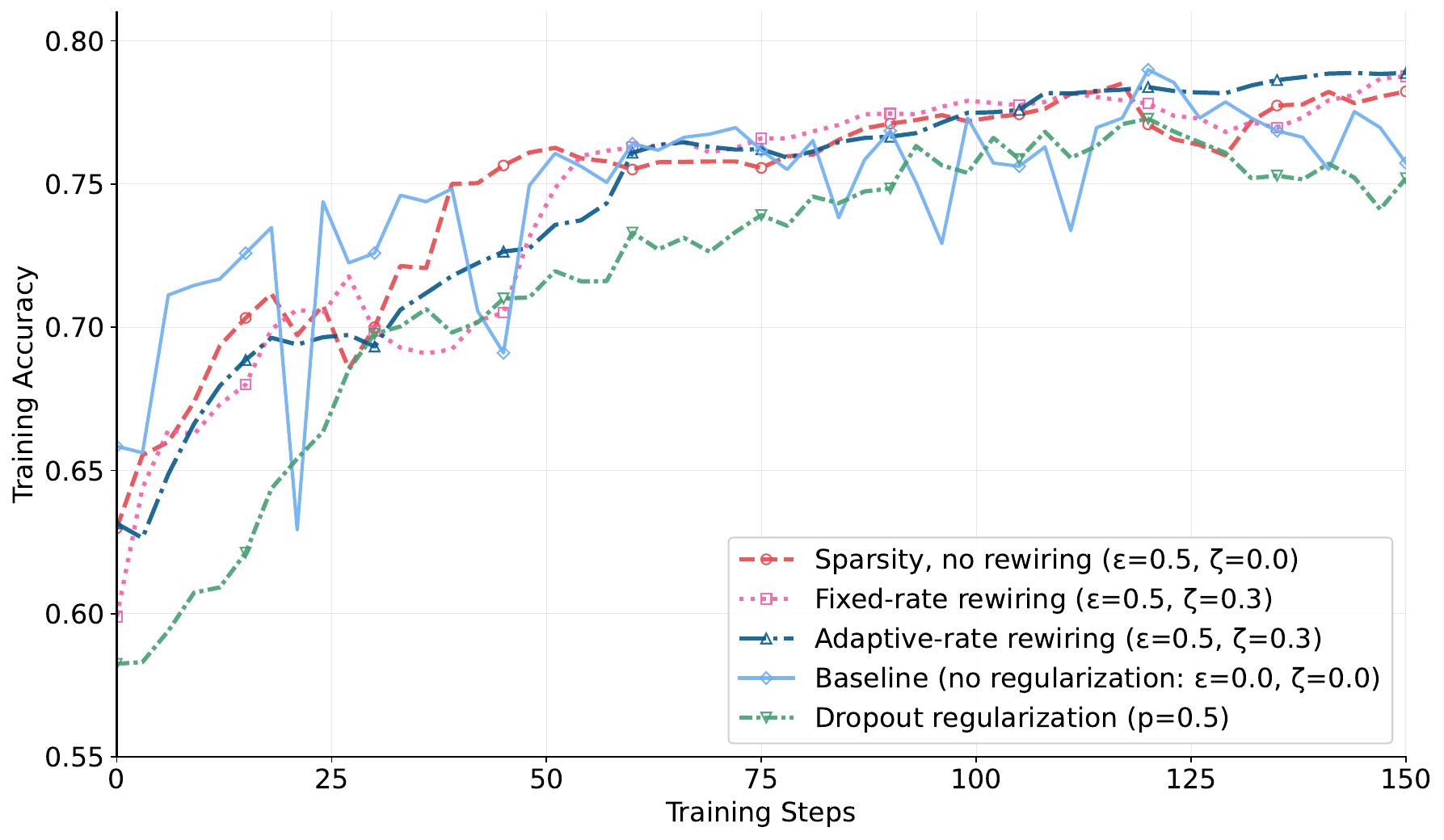}
    \caption{Training accuracy of different regularization techniques for the GIN model on the PROTEINS dataset, showing the performance of different sparsity and rewiring configurations: sparsity ($\epsilon=0.5$), fixed-rate rewiring ($\zeta_f=0.3$), adaptive-rate rewiring ($\zeta_a=0.3$), the baseline ($\epsilon=0$, $\zeta=0$), and with dropout. The results are averaged over 50 runs.}
    \label{fig:train_gin_proteins}
\end{figure*}

Figure~\ref{fig:train_gin_proteins} illustrates the training accuracy of the GIN model on the PROTEINS dataset over training steps, comparing different sparsity and rewiring strategies. The plot shows the training accuracy as a function of the training steps for four configurations:  sparsity, fixed-rate rewiring, adaptive rewiring, and baseline GIN (standard). The GIN model with sparsity achieves a comparable final accuracy to the baseline. The training dynamics of different rewiring strategies offer insight into how they function as regularizers. Adaptive rewiring, in particular, demonstrates a smoother and more stable training curve compared to fixed-rate rewiring or baseline models. This is likely because the adaptive process continuously optimizes the graph's structure alongside model weights, which provides a more robust and efficient learning trajectory. Fixed-rate rewiring demonstrated a similar performance trend to  sparsity, but achieves a slightly higher final accuracy (0.83 vs. 0.79), indicating that fixed-rate rewiring can yield a modest performance improvement. Adaptive rewiring shows a smoother trend, achieving a significantly higher final accuracy than both the fixed-rate rewiring and the baseline, reaching 0.87. 

Dropout, another common regularization technique, can be applied to GNNs as well. While rewiring modifies the graph's structure by adding or removing edges, dropout randomly deactivates nodes or edges during training. Both methods introduce randomness to prevent the model from becoming too reliant on specific connections, but they do so at different levels: rewiring at the structural level and dropout at the node, or edge, activity level. In some cases, a combination of rewiring and dropout could offer complementary regularization benefits.

In addition to the configurations shown in Figure~\ref{fig:train_gin_proteins}, similar performance trends were observed in other combinations of datasets and GNN models for the levels of sparsity and rewiring strategies. In general, all configurations converge to a similar range of final training accuracy. This suggests that while different sparsity and rewiring strategies affect the training dynamics of the GIN model, they result in comparable final training accuracies on the PROTEINS dataset, with adaptive rewiring offering the best overall performance.

\subsection{Trade-off analysis between model sparsity and rewiring}
To further understand the interplay of sparsity parameters, we analysed the combined effect of the sparsity of the model and the adaptive (resp., fixed-rate) rewiring.

\begin{figure*}[t]
    \centering
    \includegraphics[width=0.65\linewidth]{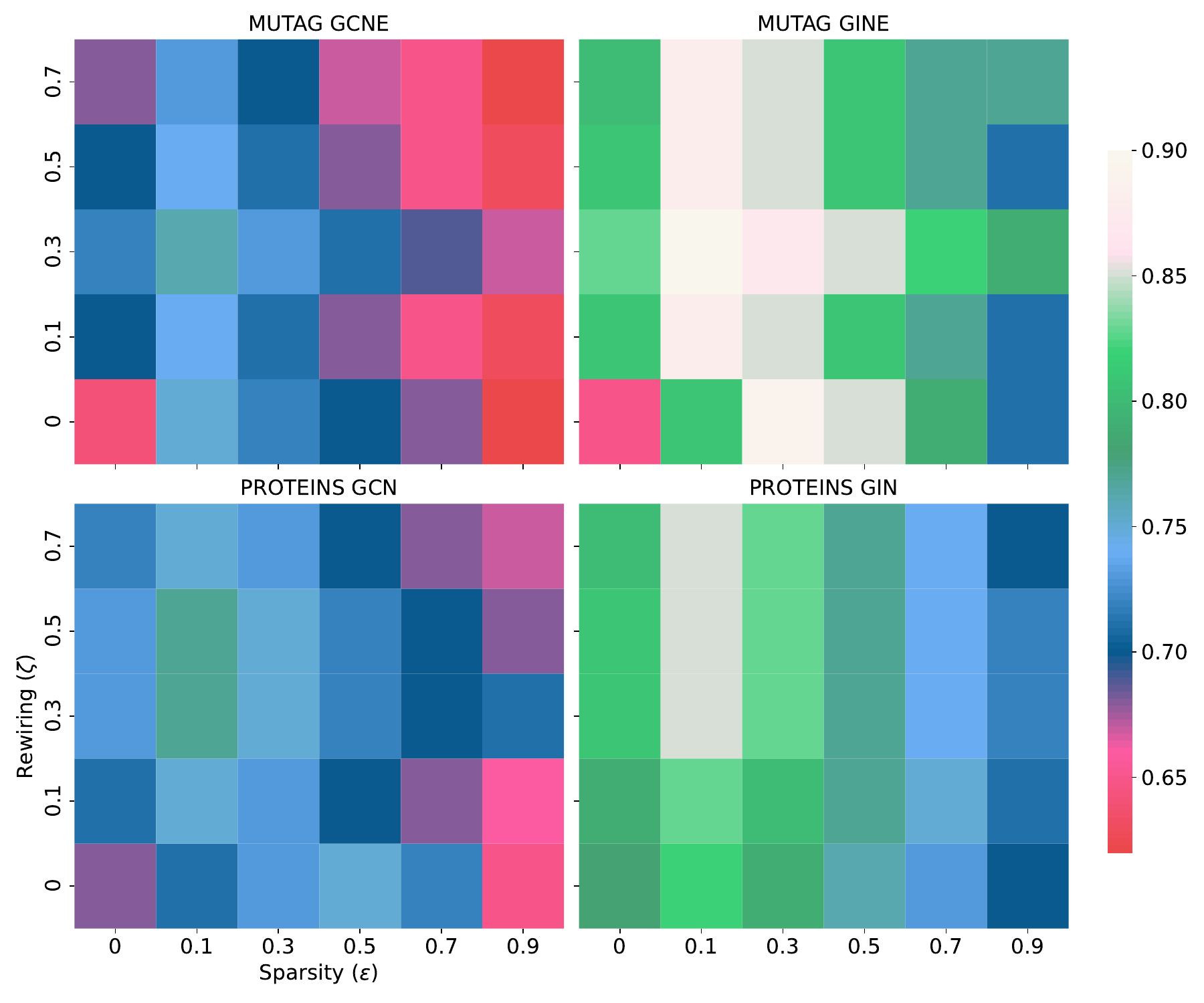}
    \caption{Heatmaps illustrating the combined impact of model sparsity ($\epsilon$), and the average $\zeta$ value between fixed-rate, and adaptive-rate rewiring on GNN performance for the MUTAG (top) and PROTEINS (bottom) datasets. Left: GCNE model. Right: GINE model. The results are averaged over 50 runs, and the colorbars indicate model test accuracy.}
    \label{fig:contour1}
\end{figure*}

Figure~\ref{fig:contour1} displays the influence of sparsity applied at the initialisation ($\epsilon$) and during rewiring (both fixed-rate and adaptive, averaging $\zeta$) on GNN accuracy across the MUTAG and PROTEINS datasets through heatmaps. 

Accuracy varies significantly with both sparsity parameters, reflecting the model's and dataset's dependence. We observe non-linear relationships, with optimal sparsity regions indicated by lighter hues. GIN models generally surpass GCN models, and dataset-specific performance patterns are observed. Colour bars provide accuracy values. GIN models might surpass GCN models due to their ability to better capture node and edge features through learnable aggregation functions, providing more expressive power.

\subsection{Number of model parameters}

\begin{table*}[ht]
    \centering
    \caption{Model Size (amount of parameters) for different datasets and sparsity levels ($\epsilon$). \label{tab:model_size_epsilon}}
    \begin{tabular}{l|l|c|c|c|c|c|c}
    \hline
    \multicolumn{1}{l|}{\textbf{Dataset}} & \multicolumn{1}{l|}{\textbf{Model}}   & \multicolumn{5}{c}{\textbf{Sparsity $\epsilon$}}  \\ \cline{3-8}
    &  & \textbf{0} & \textbf{0.1} & \textbf{0.3} & \textbf{0.5} & \textbf{0.7} & \textbf{0.9} \\
    \hline
    \rowcolor{lightgray}MUTAG & GCNE & 20 000 & 18 000 & 14 000 & 10 000 & 6 000 & 2 000 \\
    MUTAG & GINE & 21 500 & 19 350 & 15 050 & 10 750 & 6 540 & 2 150 \\
    \rowcolor{lightgray}PROTEINS & GCN & 9 000 & 8 100 & 6 300 & 4 500 & 2 700 & 900 \\
    PROTEINS & GIN & 12 000 & 10 800 & 8 400 & 6 000 & 3 600 & 1 200 \\
    \hline 
    \end{tabular}
\end{table*}

To investigate the impact of sparsity on the model size, we calculated the number of parameters for each model at different levels of sparsity, $\epsilon$. The sparsity approaches with $\zeta_f$ and $\zeta_a$ do not affect the size of the model because the sparsity levels remain constant during training. 
Table~\ref{tab:model_size_epsilon} reveals a consistent trend: as the sparsity increases, the size of our GNN models decreases proportionally across both MUTAG and PROTEINS datasets. This reduction, observed in GCNE and GINE architectures, demonstrates the effectiveness of sparsity in creating lighter and more efficient models. We also observe that models GINE and GIN have a higher parameter count than GCN and GCNE. 

\subsection{N-1 case study}\label{sec:use_case}
\begin{table*}[tb]
    \centering
    \caption{Effect of sparsity ($\epsilon$) on GNN model test accuracy and the N-1 dataset. The results are averaged over 50 runs, and the best results are highlighted in bold.\label{tab:comparison_epsilon_N-1}}
    \begin{tabular}{l|l|c|c|c|c|c|c}
    \hline
    \multicolumn{1}{l|}{\textbf{Dataset}} & \multicolumn{1}{l|}{\textbf{Model}}   & \multicolumn{5}{c}{\textbf{Sparsity $\epsilon$}}  \\ \cline{3-8}
    &  & \textbf{0} & \textbf{0.1} & \textbf{0.3} & \textbf{0.5} &\textbf{0.7} & \textbf{0.9} \\
    \hline
    \rowcolor{lightgray} N-1 & GCNE & 0.62 & 0.65 & \textbf{0.71} & 0.68 & 0.63 & 0.55 \\
    N-1 & GINE & 0.92 & 0.94 & \textbf{0.98} & 0.95 & 0.91 & 0.85 \\
    \hline
    \end{tabular}
\end{table*}

\begin{table*}[tb]
    \centering
    \caption{Effect of rewiring ($\zeta_f$ and $\zeta_a$) on GNN model test accuracy and the N-1 dataset. The results are averaged over 50 runs, and the best results per rewiring strategy are underlined, and overall over both strategies are highlighted in bold. For each $\zeta$ setting, results are averaged across sparsity levels of $\epsilon$ = 0.1, 0.3, 0.5, and 0.7. An important distinction between fixed-rate and adaptive-rate rewiring is that the values of $\zeta_{a}$ correspond to the initial rewiring intensities around which the adaptive mechanism adjust during training.\label{tab:comparison_zetas_nmin1}}
    \begin{tabular}{l|l|c|c|c|c|c|c|c|c|c|c}
    \hline
    \multicolumn{1}{l|}{\textbf{Dataset}} & \multicolumn{1}{l|}{\textbf{Model}}   & \multicolumn{5}{c|}{\textbf{Fixed rewiring $\zeta_f$}} & \multicolumn{5}{c}{\textbf{Adaptive rewiring $\zeta_a$}}  \\
    \cline{3-12}
     &  & \textbf{0} & \textbf{0.1} & \textbf{0.3} & \textbf{0.5} & \textbf{0.7} & \textbf{0} & \textbf{0.1} & \textbf{0.3} & \textbf{0.5} & \textbf{0.7} \\
    \hline
    \rowcolor{lightgray}
    N-1 & GCNE & 0.62 & 0.69 & \underline{0.73} & 0.70 & 0.66 & 0.62 & 0.70 & \textbf{\underline{0.75 }}& 0.72 & 0.69 \\
    N-1 & GINE & 0.93 & 0.95 & \underline{0.97} & 0.94 & 0.90 & 0.93 & 0.96 & 0.98 & \textbf{\underline{0.99}} & 0.96 \\
    \hline
    \end{tabular}
\end{table*}

\begin{figure*}[htb]
    \centering
    \includegraphics[width=0.675\linewidth]{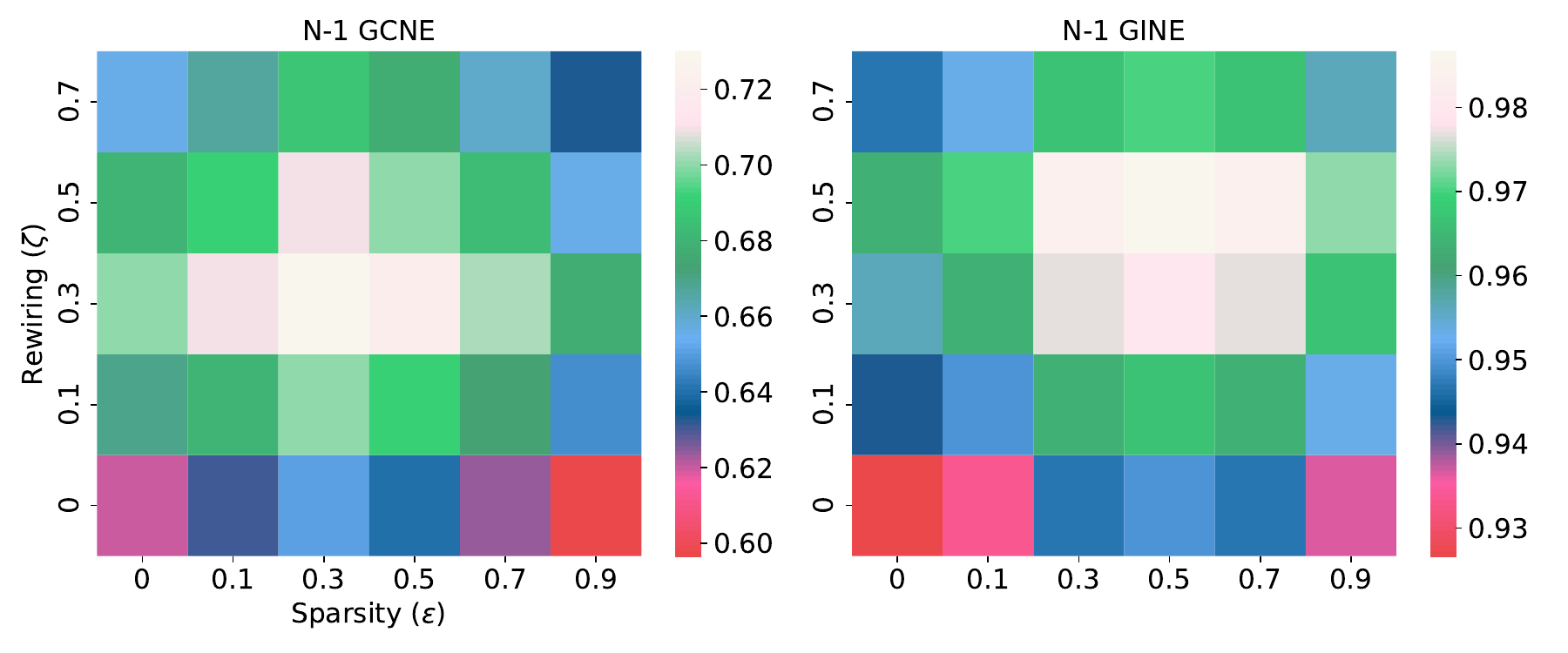}
    \caption{Heatmaps illustrating the combined impact of model sparsity ($\epsilon$), and the average $\zeta$ value between fixed-rate, and adaptive-rate rewirings on GNN performance for the N-1 dataset. Left: GCNE model. Right: GINE model. The results are averaged over 50 runs, and the colorbars indicate model accuracy. Due to performance variations, each heatmap has its own colorscale.}
    \label{fig:n1_contour_plots}
\end{figure*}

In the context of N-1 contingency analysis, the accuracy of GNN predictions directly translates into the reliability of identifying potential vulnerabilities in the energy grid. Tables~\ref{tab:comparison_epsilon_N-1} and \ref{tab:comparison_zetas_nmin1} present a comparative analysis of the performance of the GNN models on the N-1 dataset under different configurations for the model sparsity and the rewiring approaches, respectively.

Table~\ref{tab:comparison_epsilon_N-1} illustrates the effect of sparsity $\epsilon$ on the accuracy of the model. In particular, for both the GCNE and GINE models, the highest performance is achieved at an intermediate sparsity level of $\epsilon=0.3$. GCNE reaches 0.71, while GINE achieves its peak accuracy of 0.98. Performance decreases as sparsity increases beyond this point, with high sparsity ($\epsilon = 0.9$) leading to a significant drop in accuracy for both models. This suggests that while some sparsity can be beneficial, excessive sparsity has a negative impact on model learning.

Table~\ref{tab:comparison_zetas_nmin1} examines the influence of fixed-rate ($\zeta_f$) and adaptive ($\zeta_a$) rewirings on performance. It reveals that both strategies can improve accuracy. For GCNE, the best fixed-rate rewiring performance is 0.73 at $\zeta_f =0.3$, while the best adaptive rewiring performance is a slightly higher 0.75 at $\zeta_a =0.3$. For GINE, fixed-rate rewiring reaches a peak of 0.97 at $\zeta_f =0.3$, but adaptive rewiring achieves a remarkable 0.99 at a higher rate of $\zeta_a =0.5$. The data shows that while moderate rewiring is beneficial, excessively high rates (\textit{e.g.}, $\zeta_f =0.7$ and $\zeta_a =0.7$) lead to performance degradation. This is likely caused by the early stopping that occurred during the training process. 

These results suggest that adaptive rewiring, when properly tuned, offers significant advantages over fixed-rate or sparsity in improving model performance. The high accuracy, especially with the GINE model, implies a robust ability to predict the impact of component failures, which is crucial to prevent blackouts.

Figure~\ref{fig:n1_contour_plots} illustrates the combined impact of sparsity, fixed-rate, and adaptive rewiring on the performance of the GNN within the N-1 dataset, revealing different model behaviors. The GCNE architecture demonstrates sensitivity to increased sparsity, with a noticeable decline in performance across higher epsilon and $\zeta$ values, suggesting vulnerability to connection reduction. In contrast, the GINE architecture exhibits good resilience, maintaining high accuracy even with substantial sparsity. The figures provide an understanding of these performance variations, emphasizing the trade-off between model complexity and the introduced sparsity efficiency.

The observed performance variations across different sparsity configurations highlight the importance of carefully tuning sparsity parameters to optimize GNN performance for this critical task. For example, the resilience of the GINE model to higher sparsity levels suggests its potential suitability for large-scale power grid analysis, where computational efficiency is crucial. By utilising optimal sparsity configurations, we can further strengthen GNN-based N-1 contingency analysis tools, making them both accurate and computationally efficient. This enables real-time assessment of grid vulnerabilities and contributes to improved overall power system reliability.

\section{Conclusion}
Introducing controlled sparsity into Graph Neural Networks (GNNs) offers a potential solution to reduce computational costs while maintaining or even improving performance. In this context, our study focuses on investigating the impact of (and the trade-off between) model sparsity and fixed-rate (resp., adaptive) rewiring as regularization on GNN performance. This study introduces unique aspects to the field by implementing a novel sparsity approach that strategically integrates adaptive rewiring to dynamically adjust model connectivity. Unlike previous works that apply sparsification with fixed and fixed-rate pattern, our approach dynamically modifies model's connectivity during training based on the evolving model parameters.

Our experiments used standard benchmark datasets, namely MUTAG and PROTEINS, and we also introduced a case study based on a real-world power grid application: N-1 reliability analysis. This allowed us to investigate the impact of different sparsification strategies within an application that includes larger graphs with edge and node features.

The results indicate that, while sparsity can act as a regularizer and improve generalization, excessive sparsity may hinder the model's ability to learn complex patterns. Through our experiments, we observed that a sparsity level exceeding 30\% tends to degrade performance, marking it as a critical threshold where the model struggles with more complicated pattern recognition. Adaptive rewiring, particularly when combined with early stopping, proves to be a promising approach, allowing the model to adapt its connectivity structure during training and potentially leading to improved performance. This acts as a regularization mechanism that complements existing techniques like dropout by specifically targeting graphs and GNNs. For the newly introduced case study based on the N-1 dataset, we found that the GINE architecture demonstrates resilience to higher levels of sparsity, maintaining high accuracy, which is crucial for predicting the impact of component failures and preventing blackouts in power grids. This suggests that computational benefits are not compromised by a significant loss in predictive accuracy. Additionally, as we have demonstrated, high sparsity levels yield significantly smaller models with substantially fewer parameters. This reduction leads to lower memory consumption and computational requirements, making the approach more energy-efficient, environmentally friendly, and easier to scale in resource-constrained settings. For example, in a control-room scenario during an N-1 event, faster inference from a sparse GNN could enable grid operators to quickly assess potential failure points and make more informed decisions on rerouting power or initiating preventive measures, thereby enhancing operational efficiency and reducing the risk of cascading failures. As models become larger and more expensive to train, sparsity techniques provide a more sustainable alternative.

In summary, this research contributes to the understanding of how rewiring serves as a powerful form of regularization that can be effectively leveraged in GNNs for critical applications such as power grid reliability analysis. By optimizing sparsity configurations, it is possible to develop GNN-based tools that are accurate and computationally efficient, enabling real-time assessment of grid vulnerabilities and enhancing the overall resilience of the power system. 

Further research could explore adaptive methods for determining optimal sparsity levels and investigate the applicability of these techniques to other large-scale graph-based problems.

\bibliography{references_IEEE.bib}

\end{document}